\documentclass[12pt]{article}
\usepackage{graphicx}
\usepackage{hyperref}
\usepackage{comment}
\usepackage{algorithm2e}
\usepackage{color}
\usepackage{amsmath, amssymb}

\graphicspath{{.}}

\def\institution{Technical Report}
\def\support{\footnote{https://github.com/kirk86/ImageRetrieval}}

\def\Title#1{\begin{center} {\Large #1 } \end{center}}
\def\Author#1{\begin{center}{ \sc #1} \end{center}}
\def\Address#1{\begin{center}{ \it #1} \end{center}}

\newenvironment{Abstract}{\begin{quotation}  }{\end{quotation}}
\newenvironment{Presented}{\begin{quotation} \begin{center}
             PRESENTED AT\end{center}\bigskip
      \begin{center}\begin{large}}{\end{large}\end{center} \end{quotation}}
  
\def\beq{\begin{equation}}
\def\eeq#1{\label{#1}\end{equation}}
\def\eeqn{\end{equation}}

\def\beqa{\begin{eqnarray}}
\def\eeqa#1{\label{#1}\end{eqnarray}}
\def\eeqan{\end{eqnarray}}

\let\bar=\overbar

\def\Dslash{\not{\hbox{\kern-4pt $D$}}}
\def\dslash{\not{\hbox{\kern-2pt $\del$}}}

\def\msb{{\bar{\ssstyle M \kern -1pt S}}}

\begin{document}
\begin{titlepage}

\vfill
\Title{Content-based image retrieval tutorial}
\vfill
\Author{Joani Mitro}
\begin{center}
  \begin{small}
    giannismitros@gmail.com
  \end{small}
\end{center}
\Address{\institution}
\vfill
\begin{Abstract}
  This paper functions as a tutorial for individuals interested to
  enter the field of information retrieval but wouldn't know where to
  begin from. It describes two fundamental yet efficient image
  retrieval techniques, the first being \textit{k - nearest neighbours
    (knn)} and the second \textit{support vector machines (svm)}. The
  goal is to provide the reader with both the theoretical and
  practical aspects in order to acquire a better understanding. Along
  with this tutorial we have also developed the equivalent
  software\support\; using the MATLAB environment in order to illustrate
  the techniques, so that the reader can have a hands-on experience.
\end{Abstract}
\vfill
\vfill
\end{titlepage}
\def\thefootnote{\fnsymbol{footnote}}
\setcounter{footnote}{0}

\section*{Notation}

\begin{table}[h]
 \caption{Notation}
 \vspace{0.3cm}
\begin{tabular}{ll}
 \textbf{X, Y, M} & bold face roman letters indicate matrices \\
 $\vec{\mathbf{x}}, \vec{\mathbf{y}}, \vec{\mathbf{v}}$ & bold face
small letters indicate
vectors \\
  a, b, c & small letters indicate scalar values \\
 ${(\sum_{i=1}^{N}\vert x_{i}-y_{i}\vert^{p})}^{\frac{1}{p}}$ & p-norm
  distance function\\
 ${\lim_{p\rightarrow\infty}(\sum_{i=1}^{N}\vert
x_{i}-y_{i}\vert^{p})}^{\frac{1}{p}}$ & infinity norm distance function\\
  $g(z)$ & nonlinear function (e.g.\ sigmoid, tanh etc.) \\
  $z = \mathbf{W}^{T}\vec{\mathbf{x}} + \vec{\mathbf{b}}$ & score
                                                            function,
                                                            mapping
                                                            function \\
  $\Vert\vec{\mathbf{b}}\Vert_{2} = 1$ & vector norm \\
  $\phi(\vec{\mathbf{x}})$ & feature mapping function \\
  $K(\vec{\mathbf{x}}, \vec{\mathbf{z}})$  & Kernel function \\
  $K_{m}$ & Kernel matrix
\end{tabular}
\label{tab:notation_table}
\end{table}

\section{Introduction}

As we have already mentioned this tutorial serves as an introduction
for to the field of information retrieval for the interested
reader. Apart from that, there's always been a motivation for the
development of efficient media retrieval systems, since the new era of
digital communication has brought an explosion of multimedia data over
the internet. This trend, has continued with the increasing
popularity of imaging devices, such as digital cameras that nowdays are an
inseparable part of any smartphone, together with an inceasing
proliferation of image data over communication networks.

\section{Data pre-processing}

Like in any other case before we use our data we first have to
clean them if that is necessary and transform them into a format that
is understanble by the prediction algorithms. In this particular case
the process that has been adopted includes the following six steps,
applied for each image in our dataset $\mathcal{D}$, in order to
transform the raw pixel images into something meaningful that the
prediction algorithms can understand. In another sense, we map the raw
pixel values into a feature space.

\begin{enumerate}
    \item We start by computing the color histogram for each image. In
    this case the HSV color space has been chosen and each H, S, V
    component is uniformly quantized into 8, 2 and 2 bins
    resepctively. This produces a vector of 32 elements/values
    for each image.
    \item The next step is to compute the color auto-correlogram for
    each image, where the image is quantized into $4\times 4\times
    4 = 64$ colors in the RGB space. This process produces a vector of
    64 elements/values for each image.
    \item Next, we extract the first two moments (i.e.\;mean and
    standard deviation) for each R,G,B color channel. This gives us a
    vector of 6 elements/values.
    \item Moving forward, we compute the mean and standard deviation
    of the Gabor wavelet coefficients, which produces a vector of 48
    elements/values. This computation requires applying the Gabor
    wavelet filters for each image spanning accross four scales:
    ``0.05, 0.1, 0.2, 0.4'' and six orientations: ``$\theta_{0} = 0,
    \theta_{n+1} = \theta_{n} + \frac{6}{\pi}$''.
    \item Last but not least, we apply the wavelet transform to each
    image with a 3-level decomposition. In this case the mean and
    standard deviation of the transform coefficients is utilized to
    form the feature vector of 40 elements/values for each image.
    \item Finally, we combine all the vectors from the step 1--5 into
    a new vector $\vec{\mathbf{\rho}} = 32 + 64 + 6 + 48 + 40 +
    1$. Each number indicates the dimensionality of the vectors from
    steps 1--5 that have been concatenated into the new vector
    $\vec{\mathbf{\rho}}$.
\end{enumerate}

\section{Methodology}

\subsection{k-Nearest Neighbour}
\label{sec:k-NN}
k-Nearest Neighbour (k-NN) classifier belongs to the family of
instance based learning algorithms (IBL). IBL algorithms construct
hypothesis directly from the training data themselves which means that
the hypothesis complexity can grow with the data. One of its
advantages is the ability to adapt its model to previously unseen
data. Another advantage is the low cost of updating object instances
and also the fast learning rate since it requires no training. Some
other examples of IBL algorithms besides k-NN are kernel machines and
RBF networks. Some of the disadvantages of IBL algorithms including
k-NN, besides the computational complexity, which we already
mentioned, is the fact that they fail to produce good results with
noisy, irrelevant, nominal or missing attribute values. They also
don't provide a natural way of explaining how the data is
structured. The efficacy of k-NN algorithm relies on the use of a user
defined similarity function, for instance a p-norm distance function,
which depicts the nearest neighbours and the chosen set of
examples. It is also often used as a base procedure in benchmarking
and comparative studies. Due to the nature that it doesn't requrie any
trainnig when compared to any trained based rule, it is expected the
trained based rule to perform better, if it doesn't then the trained
base rule is deemed useless for the application under study.

Since nearest neighbour rule is a fairly simple algorihtm most
textbooks will have a short reference to it but will neglect to
provide any facts about who invented the rule in the first
place. Macello Pelillo~\cite{pelillo-alhazen-2014} tried to give an
answer to this question. Pelillo refers often to the famous Cover and
Hart paper (1967)~\cite{cover-hart} which shows what happens if a very
large selectively chosen training set is used. Before Cover and Hart
the rule was mentioned by Nilsson (1965)~\cite{nilsson} who called it
``minimum distance classifier'' and by Sebestyen
(1962)~\cite{sebestyen-review-1966}, who called it ``proximity
algorithm''. Fix and Hodges~\cite{fix-hodges} in their very early
discussion on non-parametric discrimination (1951) already pointed to
the nearest neighbour rule as well. The fundamental principle known as
Ockham's razor: ``select the hypothesis with the fewest assumption''
can be understood as the nearest neighbour rule for nominal
properties. It is, however, not formulated in terms of
observations. Ockham worked in the 14th century and Emphasized
Observations before ideas. Pelillo pointed out that this was already
done prior to Ockham, by Alhazen~\cite{alhazen} (Ibn al-Haytham), a
very early scientist (965--1040) in the field of optics and
perception. Pelillo cites some paragraphs where he shows that Alhazen
describes a training procedure as a ``universal form'' which is
completed by recalling the original objects which in this case Alhazen
refered to as particular forms.

To better understand the k-NN rule we will setup the concept of our
application. Suppose that we have a dataset comprised of 1000 images
in total, categorized in 10 different categories/classes where each one
includes 100 images.

Given an image $\textbf{I}\in\mathbf{R}^{m\times n}$ we would like to
find all possible similar images from the pool of candidate images
(i.e.\ all similar images from the dataset of 1000 total images). A
sensible first attempt algorithm would look something like this:\\

\begin{algorithm}[H]
  \KwData{$\mathcal{D} = \{image_{1}, image_{2}, image_{3},\ldots, image_{N}\}$,
  ``the set of all images''}
  \KwData{$\mathbf{I}\in\mathbf{R}^{m\times n}$, ``query image,
    you're trying to identify similarity against $D$''}
  \KwResult{$d\in\mathbf{R}$, ``a scalar indicating how similar two images
    are''}
  \For{image in $\mathcal{D}$}{
    \For{column in image height: $m\leftarrow height(\mathbf{I})$}{
      \For{row in image width: $n\leftarrow width(\mathbf{I})$}{
        $d[row][column] \leftarrow abs\left(\mathbf{I}[row][column] - image[row][column]\right)$
      }
    }
    $d[image] \leftarrow \mbox{sum}\left( d[row][column] \right)$ ``sum accros
    rows and columns''
  }
 \caption{naive k-NN algorithm.}
\label{algo:Algorithm1}
\end{algorithm}

Here is a visual representation of what it might look like:

\begin{figure}[h]
\includegraphics{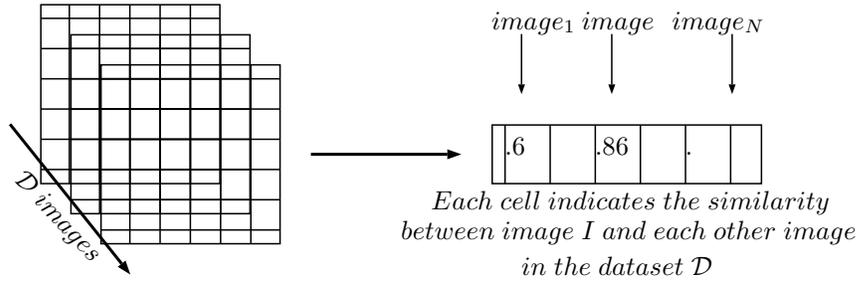}
\centering
\caption{Final visual result of the Algorithm~\ref{algo:Algorithm1}}
\end{figure}

The complexity of the naive k-NN is $O(D m n)$. Can we do
better than that? Of course we can, if we avoid some of those loops by
vectorizing our main operations. Instead of operating on the 2-D images
we can vectorize them first and then perform the operaions. First we
transform our images from 2-D matrices to 1-D vectors like it is being
demonstrated in the figure below.

\begin{figure}[h]
\includegraphics{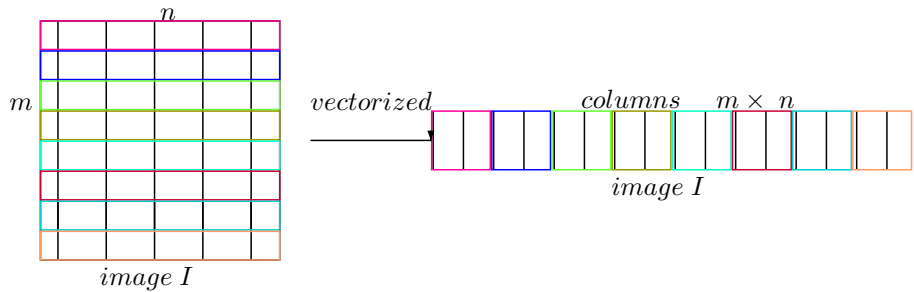}
\centering
\caption{Vectorizing images.}
\end{figure}

\begin{figure}[!h]
\includegraphics{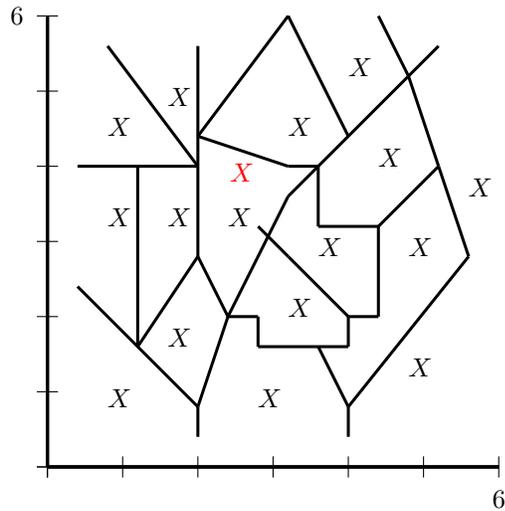}
\centering
\caption{Voronoi diagram of k-NN.}
\label{fig:voronoi}
\end{figure}

If we denote the vectorization of our query image
$I\in\mathbf{R}^{m\times n}$ as $\vec{\textbf{x}}\in\mathbf{R}^{d}$
and with $\vec{\textbf{d}}\in\mathbf{R}^{d}$ the vectorization of
every other image in the dataset $D$. Then our k-NN algorithm can be
described as follows:
$\sum_{image=1}^{N} \vert \vec{\mathbf{x}} -
\vec{\mathbf{d}}_{image}\vert$, and the complexity has now been
reduced to $O(D)$. The choice of distrance metric or distance function
is solely up to the discresion of the user. Another view of how k-NN
algorithm operates is depicted in Figure~\ref{fig:voronoi}. Notice
that k-NN performs an implicit tessellation of the feature space that
is not visible to the observer, but it is through this tessellation,
that is able to distinguish nearby datum and classify them as
similar. For instance, let's pretend that the black capital ``X''
letters in Figure~\ref{fig:voronoi} denote some data projected on the
feature space. When a new datum comes in, such as in this case the red
capital ``\textcolor{red}{X}'' letter, which indicates a query
image, then the algorithm can easily distinguish and assign to it the
closest images which are semantically similar.

\subsection{Suppoprt Vector Machines}
\begin{figure}[h]
\includegraphics{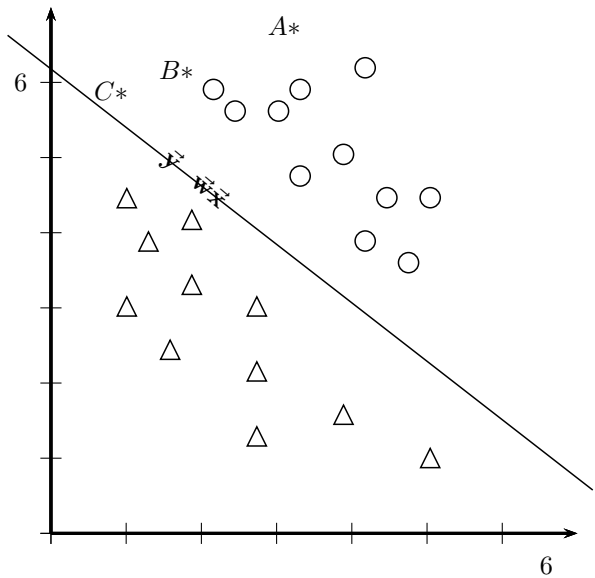}
\centering
\caption{Images projected on a 2-D plane.}
\label{fig:margins}
\end{figure}

Support Vector Machines (SVMs also known as suppport vector networks)
are supervised learnig models used among others for classifcation and
regression analysis. They were introduceds in 1992 in the Conference
on Learnning Theory by Boser, Guyon and Vapnik. It became quite
popular since then because it is a theorectically well motivated
algorithm which was developed since the 60s from Statistical Learning
Theory (Vapnik and Chervonenkis) and it also holds good empirical
performance in a diverse number of scientific fields and application
domains such as bioninformatics, text and image recognition, music
retrieval and many more. SVMs are based on the idea of separating data
with a large ``gap'' also know as margins. During the presentation of
SVM we'll also concern ourselfves with the question of optimal maring
classifier which will act as stepping stone for the introduction to
Lagrange duality. Another aspect of SVMs which is important is the
notion of kernels which allow SVMs to be applied efficiently in high
dimenisonal feature spaces. Let's start by settting up our poblem. In
this case the context is known from before where we have images from
different classes and we want to classify them accordingly. In other
words this is a binary classifcation problem. Based on this
classification we will be able to retrieve images that are similar to
our query image. Figure~\ref{fig:margins} depicts two classes of
images, the \textcolor{red}{positive}, and the
\textcolor{magenta}{negative}. For the sake of the example let's
consider the circles to be the positive and the triangles to be the
negative. We also have a hyperplane separating them as well as a three
labeled data points. Notice that point $A$ is the furthest from the
decision boundary. In order to make a prediction for the value of the
label $\vec{\mathbf{y}}_{i}$ at point $A$, one might say that in this
particular case we can be more confident that the value of the label
is going to be $\vec{\mathbf{y}}_{i} = 1$. On the other hand, point C
even though it is on the correct side of the decision boundary where
we would have predicted a label value of $\vec{\mathbf{y}}_{i} = 1$, a
small change to the decision boundary could have caused the prediction
to be negative $\vec{\mathbf{y}}_{i} = -1$. Therefore, one can say
that we're much more confident about our prediction at $A$ than at
$C$.  Point B lies in-between these two cases. One can extrapolate and
say that if a point is far from the separating hyperplane, then we
might be significantly more confident in our predictions. What we are
striving for is, given a training set, find a decision boundary that
allows us to make the correct and confident predictions (i.e.\ far
from the decision boundary) on the training examples.

Let's consider our binary classification problem where we have labels
$\vec{\textbf{y}}\in\{-1, 1\}$ and features $\vec{\textbf{x}}$. Then
our binary classifirer might look like

\begin{equation}
h_{\vec{\mathbf{w}},
  \vec{\mathbf{b}}}(\vec{\mathbf{x}}) =
g(\vec{\mathbf{w}}^{T}\vec{\mathbf{x}} + \vec{\mathbf{b}})
\begin{cases}
  g(z) = 1\;\mbox{if }\;z\;\ge\;0, \\
  g(z) = 0\;\mbox{otherwise}.
\end{cases}
\label{eqn:perceptron_formulation}
\end{equation}

Now we have to distinguish between two different notions of margin
such as \textbf{functional} and \textbf{gemometric} margin. The
\textbf{functional margin} of ($\vec{\mathbf{w}}, \vec{\mathbf{b}}$) with
respect to the training example ($\vec{\mathbf{x}}_{i}$,
$\vec{\mathbf{y}}_{i}$) is

\begin{equation}
  \hat{\vec{\boldsymbol{\gamma}}}_{i} =
  \vec{\mathbf{y}}_{i}(\vec{\mathbf{w}}^{T}\vec{\mathbf{x}} + \vec{\mathbf{b}})
\end{equation}

\begin{figure}[h]
\includegraphics{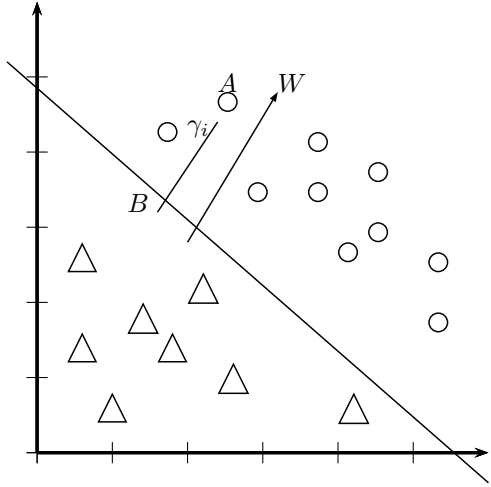}
\centering
\caption{Interpretation of geometric margin.}
\label{fig:geometric_margin}
\end{figure}

If $\vec{\mathbf{y}}_{i} = 1$, then for our prediction to be confident
and correct (i.e.\ the functional margin to be large),
$\vec{\mathbf{w}}^{T}\vec{\mathbf{x}} + \vec{\mathbf{b}}$, needs to be a
large positive number. If $\vec{\mathbf{y}}_{i} = -1$, then for the
functional margin to be large (i.e.\ to make a confident and correct
prediction) $\vec{\mathbf{w}}^{T}\vec{\mathbf{x}} + \vec{\mathbf{b}}$ needs
to be a large negative number. Note that if we replace $\vec{\mathbf{w}}$ with
$2\vec{\mathbf{w}}$ in Equation \ref{eqn:perceptron_formulation} and
$\vec{\mathbf{b}}$ with 2$\vec{\mathbf{b}}$, then since
$g(\vec{\mathbf{w}}^{T}\vec{\mathbf{x}} + \vec{\mathbf{b}}) =
g(2\vec{\mathbf{w}}^{T}\vec{\mathbf{x}} + 2\vec{\mathbf{b}})$, would not
change $h_{\vec{\mathbf{w}}, \vec{\mathbf{b}}}$ at all, which means that it
depends only on the sign, but not on the magnitude of
$\vec{\mathbf{w}}^{T}\vec{\mathbf{x}} + \vec{\mathbf{b}}$. Regarding the
\textbf{geometric margin} we'll try to interpret them using Figure
\ref{fig:geometric_margin}.

We can see the decision boundary corresponding to ($\vec{\mathbf{w}},
\vec{\mathbf{b}}$) along with the orthogonal vector
$\vec{\mathbf{w}}$. Point $A$ resembles some training example
$\vec{\mathbf{x}}_{i}$ with label $\vec{\mathbf{y}}_{i} = 1$. The
distance to the decision boundary denoted by $\gamma_{i}$ is given by
the line segment $AB$. How can we compute $\gamma_{i}$? If we consider
$\frac{\vec{\mathbf{w}}}{\Vert\vec{\mathbf{w}}\Vert}$ to be a
unit-length vector pointing in the same direction as
$\vec{\mathbf{w}}$ then point $B = \vec{\mathbf{x}}_{i} - \gamma_{i}
\cdot \frac{\vec{\mathbf{w}}}{\Vert\vec{\mathbf{w}}\Vert}$. Since this
point lies on the decision boundary then it satisfies that
$\vec{\mathbf{w}}^{T}\vec{\mathbf{x}} + \vec{\mathbf{b}} = 0$, which means
that all points $\vec{\mathbf{x}}_{i}$ on the decision boundary satisfy
the same equation. Substituting $\vec{\mathbf{x}}$ with $B$ we get
$\vec{\mathbf{w}}^{T}(\vec{\mathbf{x}}_{i} -
\gamma_{i}\cdot\frac{\vec{\mathbf{w}}}{\Vert\vec{\mathbf{w}}\Vert}) +
\vec{\mathbf{b}} = 0$. Solving for $\gamma_{i}$ we get $\gamma_{i} =
\frac{\vec{\mathbf{w}}^{T}\vec{\mathbf{x}}_{i} +
  b}{\Vert\vec{\mathbf{w}}\Vert} =
{\left(\frac{\vec{\mathbf{w}}}{\Vert\vec{\mathbf{w}}\Vert}\right)}^{T}\vec{\mathbf{x}}_{i}
+ \frac{\vec{\mathbf{b}}}{\Vert\vec{\mathbf{w}}\Vert}$. Usually the
geometric margin of ($\vec{\mathbf{w}}, \vec{\mathbf{b}}$) with respect to
the trainning example ($\vec{\mathbf{x}}_{i}, \vec{\mathbf{y}}_{i}$)
is defined to be

\begin{align}
  \vec{\boldsymbol{\gamma}}_{i} =
  \left({\left(\frac{\vec{\mathbf{w}}}{\Vert\vec{\mathbf{w}}\Vert}\right)}^{T} +
  \frac{\vec{\mathbf{b}}}{\Vert\vec{\mathbf{w}}\Vert}\right)
\end{align}

If $\Vert\vec{\mathbf{w}}\Vert = 1$, then the functional margin is
equal to the geometric margin. Notice also that the geometric margin
is invariant to rescaling of the parameters (i.e.\ if we replace
$\vec{\mathbf{w}}$ with $2\vec{\mathbf{w}}$ and $\vec{\mathbf{b}}$
with $2\vec{\mathbf{b}}$, then the geometric margin does not
change). This way it is possible to impose an arbitray scaling
contraint on $\vec{\mathbf{w}}$ without changing anything significant
from our original equation. Given a dataset $\mathcal{D} =
\{\left(\vec{\mathbf{x}}_{i}, \vec{\mathbf{y}}_{i}\right);, i =
1,\ldots,N\}$, it is also possible to define the geometric margin of
($\vec{\mathbf{w}}, \vec{\mathbf{b}}$) with respect to $\mathcal{D}$
to be the smallest of geometric margins on the individual training
examples $\gamma = \min_{i = 1,\ldots,N}
\vec{\boldsymbol{\gamma}}_{i}$. Thus, the goal for our classifier is
to find a decision boundary that maximizes the geometric margin in
order to reflect a confident and correct set of predictions, resulting
in a classifier that separates the positive and negative training
examples with a geometric margin. Supposed that our training data are
linearly separable, how do we find a separating hyperplane that
achieves the maximum geometric margin? We start by posing the
following optimisation problem

\begin{align}
  & \max_{\vec{\mathbf{\gamma}}, \vec{\mathbf{w}}, \vec{\mathbf{b}}}
  \vec{\mathbf{\gamma}} \\
  & \mbox{s.t. } \vec{\mathbf{y}}_{i}(\vec{\mathbf{w}}^{T}\vec{\mathbf{x}}_{i} +
  \vec{\mathbf{b}}) \ge \vec{\mathbf{\gamma}},\; i = 1,\ldots,N \nonumber \\
  & \Vert\vec{\mathbf{w}}\Vert = 1 \nonumber
\end{align}

The $\Vert\vec{\mathbf{w}}\Vert = 1$ constraint ensures that the
functional margin equals to the geometric margin, in this way we are
guaranteed that all the geometric margins are at least
$\vec{\mathbf{\gamma}}$. Since ``$\Vert\vec{\mathbf{w}}\Vert = 1$''
constraint is a non-convex one, and this is hard to solve instead what
we'll try to do is transform the problem into an easier one. Consider

\begin{align}
\label{eqn:max_gamma}
  & \max_{\vec{\mathbf{\gamma}}, \vec{\mathbf{w}}, \vec{\mathbf{b}}}
    \frac{\hat{\vec{\mathbf{\gamma}}}}{\Vert\vec{\mathbf{w}}\Vert} \\
  & \mbox{s.t. } \vec{\mathbf{y}}_{i}(\vec{\mathbf{w}}^{T}\vec{\mathbf{x}}_{i} +
  \vec{\mathbf{b}}) \ge \hat{\vec{\mathbf{\gamma}}},\; i = 1,\ldots,N
    \nonumber
\end{align}

Notice that we've got ridden the constraint
$\Vert\vec{\mathbf{w}}\Vert = 1$ that was making our objective
difficult and also since
$\vec{\mathbf{\gamma}} =
\frac{\hat{\vec{\mathbf{\gamma}}}}{\Vert\vec{\mathbf{w}}\Vert}$, will
provide an acceptable and correct answer. The main problem is that
still our objective function
$\frac{\hat{\vec{\mathbf{\gamma}}}}{\Vert\vec{\mathbf{w}}\Vert}$ is
non-convex, thus we still have to keep searching for a different
representation. Recall that we can add an arbitary scaling constraint
on $\vec{\mathbf{w}}$ and $\vec{\mathbf{b}}$ without changing anything
from our original formulation. We'll introduce the scaling constraint
such that the functional margin of
$\vec{\mathbf{w}}, \vec{\mathbf{b}}$ with respect to the training set
($\vec{\mathbf{x}}_{i}, \vec{\mathbf{y}}_{i}$) must be 1, this is
$\hat{\vec{\mathbf{\gamma}}} = 1$. Multiplying
$\vec{\mathbf{w}}, \vec{\mathbf{b}}$ by some constant yields the
functional margin multiplied by the same constant. One can satisfy the
scaling constraint by rescaling
($\vec{\mathbf{w}}, \vec{\mathbf{b}}$). If we plug this consraint into
Equation~\ref{eqn:max_gamma} and substitute
$\frac{\hat{\vec{\mathbf{\gamma}}}}{\Vert\vec{\mathbf{w}}\Vert} =
\frac{1}{\Vert\vec{\mathbf{w}}\Vert}$, then we get the following
optimization problem.

\begin{align}
  \label{eqn:dual}
  & \min_{\vec{\mathbf{\gamma}}, \vec{\mathbf{w}}, \vec{\mathbf{b}}}
    \frac{1}{2} {\Vert\vec{\mathbf{w}}\Vert}^{2} \\
  & \mbox{s.t. } \vec{\mathbf{y}}_{i}(\vec{\mathbf{w}}^{T}\vec{\mathbf{x}}_{i} +
  \vec{\mathbf{b}}) \ge 1,\; i = 1,\ldots,N
    \nonumber
\end{align}

Notice that maximizing
$\frac{\hat{\vec{\mathbf{\gamma}}}}{\Vert\vec{\mathbf{w}}\Vert} =
\frac{1}{\Vert\vec{\mathbf{w}}\Vert}$ is the same thing as minimizing
${\Vert\vec{\mathbf{w}}\Vert}^{2}$. We have now transformed our
optimization problem into one with a convex quadratic objective and
linear constraints which can be solved using quadratic
programming. The solution to the above optimization problem will give
us the optimal margin classifier which will lead to the dual form of
our optimization problem, which in return plays an important role in
the use of kernels to get optimal margin classifiers, in order to work
efficiently in very high dimensional spaces. We can reexpress the
constraints of Equation~\ref{eqn:dual} as $g_{i}(\vec{\mathbf{w}}) =
-\vec{\mathbf{y}}_{i}(\vec{\mathbf{w}}^{T}\vec{\mathbf{x}}_{i} +
\vec{\mathbf{b}}) + 1 \le 0$. Notice that constraints that hold with
equality, $g_{i}(\vec{\mathbf{w}}) = 0$, correspond to training
examples $(\vec{\mathbf{x}}_{i}, \vec{\mathbf{y}}_{i})$, that have
functional margin equal to one. Let's have a look at the figure below.

\begin{figure}[h]
\includegraphics{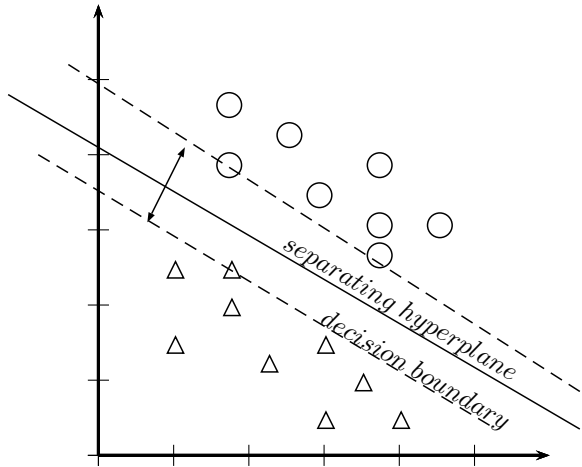}
\centering
\caption{Support vectors and maximum maring separating hyperplane.}
\label{fig:support_vectors}
\end{figure}

The three points that lie on the decision boundary (two positive and
one negative) are the ones with the smallest margins and thus closest
to the decision boundary. Notice that these three points are called
\textbf{support vectors} and usually they can be smaller in number
than the training set. In order to tackle the problem we frame it as a
Lagrangian optimization problem

\begin{align}
  \label{eqn:Lagrange_dual}
  \mathcal{L}(\alpha, \vec{\mathbf{w}}, \vec{\mathbf{b}}) =
  \frac{1}{2} {\Vert\vec{\mathbf{w}}\Vert}^{2} - \sum_{i}^{N}
  \alpha_{i}\left[ \vec{\mathbf{y}}_{i} \left( \vec{\mathbf{w}}^{T}
  \vec{\mathbf{x}}_{i} + \vec{\mathbf{b}} \right) - 1 \right].
\end{align}

with only one Lagrangian mulptiplier
``$\alpha_{i}$'' since the problem has only inequality constraints
and not any equality constraints. First, we have to find the dual form
of the problem, to do so we need to minimize $\mathcal{L}(\alpha,
\vec{\mathbf{w}}, \vec{\mathbf{b}})$ with respect to
$\vec{\mathbf{w}}$ and $\vec{\mathbf{b}}$ for a fixed
$\alpha$. Setting the derivatives of $\mathcal{L}$ with respect to
$\vec{\mathbf{w}}$ and $\vec{\mathbf{b}}$ to zero, we get:

\begin{align}
  \nabla_{\vec{\mathbf{w}}}\mathcal{L}(\alpha, \vec{\mathbf{w}},
  \vec{\mathbf{b}}) &= \vec{\mathbf{w}} - \sum_{i = 1}^{N}
  \alpha_{i}\vec{\mathbf{y}}_{i}\vec{\mathbf{x}}_{i} = 0. \\
  \vec{\mathbf{w}} &= \sum_{i = 1}^{N}
  \alpha_{i}\vec{\mathbf{y}}_{i}\vec{\mathbf{x}}_{i}.
  \begin{cases}
    \label{eqn:derivative_wrt_w}
    \mbox{derivative of }\frac{\partial{\mathcal{L}}}{\partial{\vec{\mathbf{w}}}}
  \end{cases} \\
  \frac{\partial{\mathcal{L}}(\alpha, \vec{\mathbf{w}},
  \vec{\mathbf{b}})}{\partial{\vec{\mathbf{b}}}} &= \sum_{i = 1}^{N}
  \alpha_{i}\vec{\mathbf{y}}_{i} = 0.
  \begin{cases}
    \label{eqn:derivative_wrt_b}
    \mbox{derivative of }\frac{\partial{\mathcal{L}}}{\partial{\vec{\mathbf{b}}}}
  \end{cases}
\end{align}

Substituting Equation~\ref{eqn:derivative_wrt_w} into
Equation~\ref{eqn:derivative_wrt_b} and simplifying we get

\begin{align}
  \mathcal{L}(\alpha, \vec{\mathbf{w}}, \vec{\mathbf{b}}) &=
  \sum_{i = 1}^{N} \alpha_{i} - \frac{1}{2} \sum_{i, j = 1}^{N}
  \vec{\mathbf{y}}_{i} \vec{\mathbf{y}}_{j} \alpha_{i} \alpha_{j}
  {\left( \vec{\mathbf{x}}_{i} \right)}^{T}
  \vec{\mathbf{x}}_{j} - \vec{\mathbf{b}} \sum_{i = 1}^{N} \alpha_{i}
  \vec{\mathbf{y}}_{i}. \\
  \mathcal{L}(\alpha, \vec{\mathbf{w}}, \vec{\mathbf{b}}) &=
  \sum_{i = 1}^{N} \alpha_{i} - \frac{1}{2} \sum_{i, j = 1}^{N}
  \vec{\mathbf{y}}_{i} \vec{\mathbf{y}}_{j} \alpha_{i} \alpha_{j}
  {\left( \vec{\mathbf{x}}_{i} \right)}^{T}
  \vec{\mathbf{x}}_{j}.
  \begin{cases}
    \mbox{from eq.~\ref{eqn:derivative_wrt_b} last term = 0}
  \end{cases}
\end{align}

Utilizing the constraint $\alpha_{i} \ge 0$ and the constraint from
Equation~\ref{eqn:derivative_wrt_b} the following dual optimization
problem arises:

\begin{align}
  \max_{\alpha} W(\alpha) &= \sum_{i = 1}^{N} \alpha_{i} - \frac{1}{2}
  \vec{\mathbf{y}}_{i} \vec{\mathbf{y}}_{j} \alpha_{i}\alpha_{j}
  \left<\vec{\mathbf{x}}_{i}, \vec{\mathbf{x}}_{j}\right> \\
  & \mbox{s.t. } \alpha_{i} \ge 0, i = 1,\dots,N \nonumber \\
  & \sum_{i = 1}^{N} \alpha_{i} \vec{\mathbf{y}}_{i} = 0. \nonumber
\end{align}

If we are able to solve the dual problem, in other words find the
$\alpha$ that maximizes $\vec{\mathbf{w}}(\alpha)$ then we can use
Equation~\ref{eqn:derivative_wrt_w} in order to find the optimal
$\vec{\mathbf{w}}$ as a function of $\alpha$. Once we have found
the optimal $\vec{\mathbf{w}}*$, considering the primal problem then
we can also find the optimal value for the intercept term $\vec{\mathbf{b}}$.

\begin{align}
  \vec{\mathbf{b}}* = -\frac{\max_{i:\vec{\mathbf{y}}_{i} = -1}
  {\vec{\mathbf{w}}*}^{T} \vec{\mathbf{x}}_{i} + \min_{i:
  \vec{\mathbf{y}}_{i} = 1} {\vec{\mathbf{w}}*}^{T} \vec{\mathbf{x}}_{i}}{2}
\end{align}

Suppose we've fit the parameters of our model to a training set, and we
wish to make a prediction at a new point input $\vec{\mathbf{x}}$. We
would then calculate
$\vec{\mathbf{w}}^{T}\vec{\mathbf{x}} + \vec{\mathbf{b}}$, and predict
$\vec{\mathbf{y}} = 1$ if and only if this quantity is bigger than
zero. But using Equation~\ref{eqn:derivative_wrt_w}, this quantity can
also be written:

\begin{align}
  \vec{\mathbf{w}}^{T} \vec{\mathbf{x}} + \vec{\mathbf{b}} &= {\left(
  \sum_{i = 1}^{N} \alpha_{i} \vec{\mathbf{y}}_{i}
  \vec{\mathbf{x}}_{i} \right)}^{T} \vec{\mathbf{x}} +
  \vec{\mathbf{b}} \\
  \label{eqn:dual_form2}
  & = \sum_{i = 1}^{N} \alpha_{i} \vec{\mathbf{y}}_{i} \left<
  \vec{\mathbf{x}}_{i}, \vec{\mathbf{x}} \right> + \vec{\mathbf{b}}
\end{align}

Earlier we saw that the different values for $\alpha_{i}$ will all be
zero except for the support vectors. Many of the terms in the sum
above will be zero. We really need to find only the inner products
between $\vec{\mathbf{x}}$ and the support vectors in order to
calculate Equation~\ref{eqn:dual_form2} and make our prediction. We
will exploit this property of using inner products between input
feature vectors in order to apply kernels to our classification
problem. To talk about \textbf{kernels} we'll have to think about our
input data. In this case as we have already previously mentioned we
are referring to images, and images are usually discribed by a number
of pixel values which we'll refer to as \textbf{attributes},
indicating the different intensity colors across the three different
color channels \{\textcolor{red}{R}, \textcolor{green}{G},
\textcolor{blue}{B}\}. When we process the pixel values in order to
retrieve more meaningful representations, in other words when we map our
initial pixel values through some processing operation to some new
values, these new values are called \textbf{features} and the
operation process is referred to as \textbf{feature mapping} usually
denoted as $\phi$.

Instead of applying SVM directly to the attributes
$\vec{\mathbf{x}}$, we may want to use SVM to learn from some features
$\phi({\vec{\mathbf{x}}_{i}})$. Since the SVM algorithm can be written
entirely in terms of innner products
$\left<\vec{\mathbf{x}}, \vec{\mathbf{z}}\right>$ we can instead
replace them with $\left<\phi(\vec{\mathbf{x}}),
  \phi(\vec{\mathbf{z}})\right>$. This way given a feature mapping
$\phi$, the corresponding \textbf{kernel} is defined as
$K(\vec{\mathbf{x}}, \vec{\mathbf{z}}) = {\phi(\vec{\mathbf{x}})}^{T}
\phi(\vec{\mathbf{z}})$. If we replace every inner product
$\left<\vec{\mathbf{x}}, \vec{\mathbf{z}}\right>$ in the algorithm
with $K(\vec{\mathbf{x}}, \vec{\mathbf{z}})$,
then the learning process will be happening uisng features $\phi$.

One can compute $K(\vec{\mathbf{x}}, \vec{\mathbf{z}})$ by finding
$\phi(\vec{\mathbf{x}})$ and $\phi(\vec{\mathbf{z}})$ even though they
may be expensive to calculate because of their high
dimensionality. Kernels such as
$K(\vec{\mathbf{x}}, \vec{\mathbf{z}})$, allows SVMs to perform
learning in high dimensional feature spaces without the need to
explicitly find or represent vectors $\phi(\vec{\mathbf{x}})$. For
instance, suppose
$\vec{\mathbf{x}}, \vec{\mathbf{z}}\in \mathbb{R}^{n}$, and let's
consider $K(\vec{\mathbf{x}}, \vec{\mathbf{z}}) =
{(\vec{\mathbf{x}}^{T} \vec{\mathbf{z}})}^{2}$ which is equivalent to

\begin{align}
  K(\vec{\mathbf{x}}, \vec{\mathbf{z}}) &= \left( \sum_{i = 1}^{n}
  \vec{\mathbf{x}}_{i} \vec{\mathbf{z}}_{i}
  \right) \left( \sum_{j = 1}^{n} \vec{\mathbf{x}}_{j}
  \vec{\mathbf{z}}_{j} \right) \\
  & = \sum_{i = 1}^{n} \sum_{j = 1}^{n}
  \vec{\mathbf{x}}_{i} \vec{\mathbf{x}}_{j} \vec{\mathbf{z}}_{i}
    \vec{\mathbf{z}}_{j} \nonumber \\
  & = \sum_{i, j = 1}^{n}
  \left(  \vec{\mathbf{x}}_{i} \vec{\mathbf{x}}_{j} \right) \left(
    \vec{\mathbf{z}}_{i}
    \vec{\mathbf{z}}_{j} \right) \nonumber \\
  & = {\phi(\vec{\mathbf{x}})}^{T} \phi(\vec{\mathbf{z}}) \nonumber
\end{align}

for $n = 3$ the feature mapping $\phi$ is computed as

\begin{align}
  \phi(\vec{\mathbf{x}}) &= \begin{bmatrix}
                             x_{1} x_{1} \\
                             x_{1} x_{2} \\
                             x_{1} x_{3} \\
                             x_{2} x_{1} \\
                             x_{2} x_{2} \\
                             x_{2} x_{3} \\
                             x_{3} x_{1} \\
                             x_{3} x_{2} \\
                             x_{3} x_{3} \\
                             \end{bmatrix} \nonumber
\end{align}

Broadly speaking a kernel
$K(\vec{\mathbf{x}}, \vec{\mathbf{z}}) = {(\vec{\mathbf{x}}^{T}
  \vec{\mathbf{z}} + c)}^{d}$ corresponds to a feature mapping of
$\binom{n+d}{d}$ feature space.
$K(\vec{\mathbf{x}}, \vec{\mathbf{z}})$ still takes $O(n)$ time even
though it is operating in a $O(n^{d})$ space, because it doesn't need
to explicitly represent feature vectors in this high dimensional
space. If we think of $K(\vec{\mathbf{x}}, \vec{\mathbf{z}})$ as some
measurement of how similar are $\phi(\vec{\mathbf{x}})$ and
$\phi(\vec{\mathbf{z}})$, or $\vec{\mathbf{x}}$ and
$\vec{\mathbf{z}}$, then we might expect $K(\vec{\mathbf{x}},
\vec{\mathbf{z}}) = {\phi(\vec{\mathbf{x}})}^{T}
\phi(\vec{\mathbf{z}})$ to be large if $\phi(\vec{\mathbf{x}})$ and
$\phi(\vec{\mathbf{z}})$ are close together and vice versa.

Suppose that for some learning problem we have thought of some
kernel function $K(\vec{\mathbf{x}}, \vec{\mathbf{z}})$, considered as
a reasonable measure of how similar $\vec{\mathbf{x}}$ and
$\vec{\mathbf{z}}$ are. For instace,

\begin{align}
  K(\vec{\mathbf{x}}, \vec{\mathbf{z}}) =
  {e}^{ -\frac{ {\Vert\vec{\mathbf{x}} -
  \vec{\mathbf{z}}\Vert}^{2}} {2\sigma^{2}} }
  \begin{cases}
    1\mbox{ if } \vec{\mathbf{x}} \mbox{ and } \vec{\mathbf{z}} \mbox{
      is
    close} \\
    0 \mbox{ otherwise }
  \end{cases}
\end{align}

the question then becomes, can we use this definition as the kernel in
an SVM algorithm? In general, given any function $K$ is there any
process which will allow to describe if it exists some feature mapping
$\phi$ so that $K(\vec{\mathbf{x}}, \vec{\mathbf{z}}) =
{\phi(\vec{\mathbf{x}})}^{T} \phi(\vec{\mathbf{z}})$ for all
$\vec{\mathbf{x}}$ and $\vec{\mathbf{z}}$, in other words is it a
valid kernel or not? If we suppose that $K$ is a valid kernel then
${\phi(\vec{\mathbf{x}}_{i})}^{T} \phi(\vec{\mathbf{x}}_{j}) =
{\phi(\vec{\mathbf{x}}_{j})}^{T} \phi(\vec{\mathbf{x}}_{i})$, meaning
that the kernel matrix denoted as $K_{m}$, discribing similarity
between datum $\vec{\mathbf{x}}_{i}$ and $\vec{\mathbf{x}}_{j}$, must
be symmetric. If we denote $\phi_{k}(\vec{\mathbf{x}})$, the $k$-the
coordinate of the vector $\phi(\vec{\mathbf{x}})$, then for any vector
$\vec{\mathbf{z}}$, we have

\begin{align}
  \vec{\mathbf{z}}^{T} K \vec{\mathbf{z}}
  & = \sum_{i}\sum_{j} \vec{\mathbf{z}}_{i} K_{ij} \vec{\mathbf{z}}_{j} \\
  & =  \sum_{i}\sum_{j} \vec{\mathbf{z}}_{i}
    {\phi_{k}(\vec{\mathbf{x}}_{i})}^{T}
    \phi_{k}(\vec{\mathbf{x}}_{j}) \vec{\mathbf{z}}_{j} \nonumber \\
  & =  \sum_{i}\sum_{j} \vec{\mathbf{z}}_{i}\sum_{k}
    {\phi_{k}(\vec{\mathbf{x}}_{i})} \phi_{k}(\vec{\mathbf{x}}_{j})
    \vec{\mathbf{z}}_{j} \nonumber \\
  & =  \sum_{k}\sum_{i}\sum_{j}\vec{\mathbf{z}}_{i}
    {\phi_{k}(\vec{\mathbf{x}}_{i})} \phi_{k}(\vec{\mathbf{x}}_{j})
    \vec{\mathbf{z}}_{j} \nonumber \\
  & =  \sum_{k} {\left( \sum_{i}\vec{\mathbf{z}}_{i}
    {\phi_{k}(\vec{\mathbf{x}}_{i})} \right)}^{2} \nonumber \\
  & \ge 0. \nonumber
\end{align}

which shows that the kernel matrix $K_{m}$ is positive semi-definite
($K \ge 0$) since our choice of $\vec{\mathbf{z}}$ was arbitary. If
$K$ is a valid kernel meaning that it corresponds to some feature
mapping $\phi$, then the corresponding kernel matrix
$K_{m}\in\mathbb{R}^{m\times m}$ is symmetric positive
semi-definite. This is a necessary and sufficient condition for
$K_{m}$ to be a valid kernel also called the Mercer kernel.
The take away message is that if you have any algorithm that
you can write in terms of only inner products
$\left<\vec{\mathbf{x}}, \vec{\mathbf{z}}\right>$ between the input
attribute vectors, then by replacing it with a kernel
$K(\vec{\mathbf{x}}, \vec{\mathbf{z}})$ you can permit your algorithm
to work efficiently in the high dimensional feature space.

Switching gears for a moment and returning back to our problem or
actually classifying and semantically retrieving similar images, since
we now have an understanding of how the SVM algorithm is functioning,
we can use it in our application. Recall that we have the following
dataset:

\begin{align}
\mathcal{D}_{10-\mbox{classes}} =
\begin{cases}
  \mbox{Africa} \\
  \mbox{Monum.} \\
  \mbox{Animals} \\
  \mbox{People} \\
  \hspace{0.4cm}\vdots
\end{cases} \nonumber
\leftarrow
\begin{cases}
  \begin{array}{ |c|c|c|c|c| }
    \hline
    \mbox{Africa} & \mbox{Monuments} & \mbox{Animals} & \mbox{People} & \cdots \\
    \hline
    image_{1} & image_{1} & image_{1} & image_{1} & \vdots \\
    image_{2} & image_{2} & image_{2} & image_{2} & \\
    image_{3} & image_{3} & image_{3} & image_{3} & \\
    \vdots   & \vdots    & \vdots    & \vdots   & \\
    image_{100} & image_{100} & image_{100} & image_{100} & \\
    \hline
  \end{array}
\end{cases}
\end{align}

As in Section~\ref{sec:k-NN} we have our dataset and our query image,
in this case denoted by $\vec{\mathbf{q}}$, vectorized, in order to
perform mathematical operations seamelessly. To make things even more
explicit, imagine that our system (i.e.\ the MATLAB software
accompanying this tutorial) or the algorithm (i.e.\ the SVM in this
case) receives a query image $\vec{\mathbf{q}}$ from the user and its
job is to find and return to the user all the images which are similar
to the query $\vec{\mathbf{q}}$. For instance, if the query image
$\vec{\mathbf{q}}$ of the user depicts a monument then the job of our
system or algorithm is to return to the user all the images depicting
monuments from our dataset $\mathcal{D}$.

In other words we are
treating our problem as a multiclass classification problem. Generally
speaking there are two broad approaches in which we can resolve this
issue using the SVM algorithm. The first one is called ``one-vs-all''
approach and involves training a single classifier per class, with
the samples of that class as positive samples and all other samples as
negatives. This strategy requires the base classifiers to produce a
real-valued confidence score for its decision, rather than just a
class label.

The second approach is called ``one-vs-one'' and usually
one has to train $\frac{n!}{(n-k)!k!}$ binary classifiers for a k-way
multiclass problem. Each receives the samples of a pair of classes
from the original training set, and must learn to distinguish these
two classes. At prediction time, a voting scheme is applied: all
$\frac{n!}{(n-l)!k!}$ classifiers are applied to an unseen sample and
the class that got the highest number of ``+1'' predictions gets
predicted by the combined classifier. This is the method that the
accompanying software is utilizing for the SVM solution.

Notice that if the SVM algorithm predicts the wrong class label for a
query image $\vec{\mathbf{q}}$, then we end up retrieving and returning
to the user all the images from the wrong category/class since we
predicted the wrong label. How can we compensate for this shortcoming?
This is left as an exercise to the reader to practice his/her skills.


\begin{thebibliography}{99}
\bibitem{alhazen}
Ibn al-Ḥasan, widely considered to be one fo the first theoretical
physicists, c.965--c.1040 CE
\end{thebibliography}
\end{document}